\documentclass[10pt, conference, compsocconf]{IEEEtran}
\usepackage[pdftex]{graphicx}
\usepackage{textcomp}
\usepackage{makecell}

\makeatletter
\newcommand*{\rom}[1]{\expandafter\@slowromancap\romannumeral #1@}
\makeatother

\begin{document}
\title{TimeConvNets: A Deep Time Windowed Convolution Neural Network Design for Real-time Video Facial Expression Recognition}


\author{\IEEEauthorblockN{James Ren Hou Lee and Alexander Wong}
\IEEEauthorblockA{Department of Systems Design Engineering\\
Waterloo Artificial Intelligence Institute\\
University of Waterloo\\
Waterloo, Ontario\\
jrhlee, a28wong@uwaterloo.ca}
}


%Telephone: (800) 555--1212, Fax: (888) 555--1212}

\maketitle

\begin{abstract}

A core challenge faced by the majority of individuals with Autism Spectrum Disorder (ASD) is an impaired ability to infer other people's emotions based on their facial expressions. With significant recent advances in machine learning, one potential approach to leveraging technology to assist such individuals to better recognize facial expressions and reduce the risk of possible loneliness and depression due to social isolation is the design of computer vision-driven facial expression recognition systems. Motivated by this social need as well as the low latency requirement of such systems, this study explores a novel deep time windowed convolutional neural network design (TimeConvNets) for the purpose of real-time video facial expression recognition. More specifically, we explore an efficient convolutional deep neural network design for spatiotemporal encoding of time windowed video frame sub-sequences and study the respective balance between speed and accuracy. Furthermore, to evaluate the proposed TimeConvNet design, we introduce a more difficult dataset called BigFaceX, composed of a modified aggregation of the extended Cohn-Kanade (CK+), BAUM-1, and the eNTERFACE public datasets. Different variants of the proposed TimeConvNet design with different backbone network architectures were evaluated using BigFaceX alongside other network designs for capturing spatiotemporal information, and experimental results demonstrate that TimeConvNets can better capture the transient nuances of facial expressions and boost classification accuracy while maintaining a low inference time.

\end{abstract}


% \fi
\IEEEpeerreviewmaketitle

\section{Introduction}

The ability to detect, recognize, and understand different facial expressions is a crucial skill to have in everyday life, due to the abundance of social interactions one faces on a daily basis. This skill not only grants understanding of current emotional states, but also allows the user to recognize conversational cues such as level of interest, speaking turns, and level of information understanding \cite{michel-svm}. Research has shown that a staggering 55$\%$ of the information behind a spoken message stems from facial cues, and only 7$\%$ is attributed to the words themselves \cite{lstm}. Clearly, the visual component of the message is key when interpreting a message, even more important than the aural aspect, which only takes up 38$\%$ of the spoken message \cite{lstm}. 

For the majority of individuals with Autism Spectrum Disorder (ASD), a core challenge that they face is an impaired ability to infer other people's emotions based on their facial expressions. As such, the complexity of human societal interaction is further elevated, and can often lead to a higher prevalence of loneliness and depression, due to difficulties interacting with society. Motivated by this important social need as well as the significant recent advances in machine learning, one possible approach to explore for assisting individuals with ASD to better recognize facial expressions is the design of computer vision-driven facial expression recognition systems for enabling improved emotion inference and improving social interactions.

\begin{figure}[t]
\includegraphics[width=0.5\textwidth]{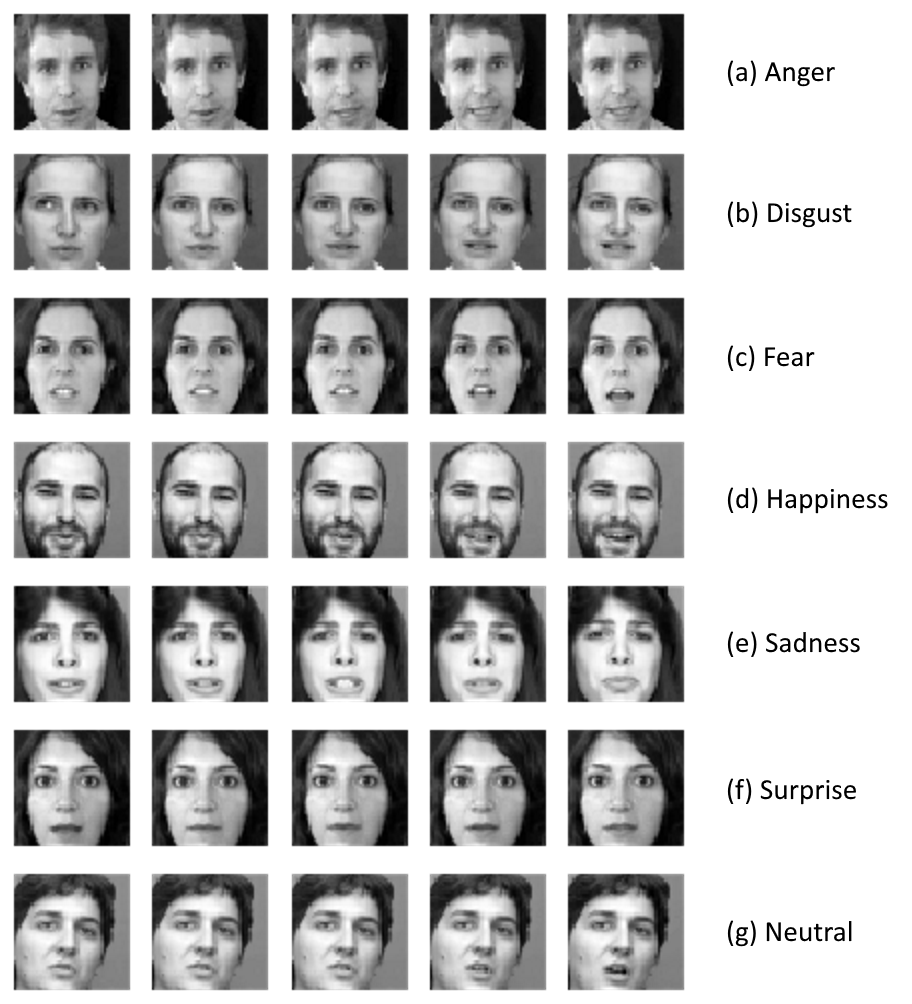}
\caption{Example facial expression data in BigFaceX for the six key universal facial expression categories, plus neutral. Each row (left to right) represents a single 5-channel sub-sequence stack of frames. (a) Anger. (b) Disgust. (c) Fear. (d) Happiness. (e) Sadness. (f) Surprise. (g) Neutral.}
\centering
\label{fig:faces}
\end{figure}

The field of facial emotion classification (FEC) has been an active area and has resulted in a wide variety of approaches over the years.  Traditional machine learning-driven strategies for FEC typically involve the coupling of a machine learning algorithm (e.g., Support Vector Machine (SVM)~\cite{michel-svm}) with a combination of different hand-engineered features and other computer vision techniques (e.g., Local Binary Patterns (LBP) features, feature point tracking, and dense optical flow \cite{bargal2016emotion}). More recently, much of the focus and advances in FEC has evolved around deep learning, with a number of different deep neural network architectures for tackling this problem based on convolutional neural networks (CNNs), recurrent neural networks (RNNs) \cite{CNN-RNN}, long short term memory (LSTM) networks \cite{lstm}, and deep dense feedforward networks \cite{3dc}.  The FEC methods proposed in research literature generally fall under one of two categories: 
\begin{enumerate}
    \item \textbf{Static single image classification}: A single image is used by the machine learning algorithm to predict a single emotion label (e.g., (anger, disgust, fear, happiness, sadness, surprise) \cite{michel-svm, 3dc}, or neutral). This category of strategies have faster inference speed but do not better capture spatiotemporal characteristics associated with facial expressions. 
    \item \textbf{Dynamic video classification}: An entire video sequence is used by the machine learning algorithm to predict a single emotion label.  This category of strategies better capture spatiotemporal characteristics at the expense of computational complexity.
    \end{enumerate}

Despite the advancements made in the field of facial emotion classification, there remains a number of challenges that need to be tackled for the purpose of helping individuals with ASD to better recognize facial expressions during social interactions:
\begin{itemize}
    \item \textbf{Low Latency Requirement}: For fluent social interactions with others, it is required that the latency between the facial expression being made and the recognition and visualization of the emotion being conveyed be as low as possible, such that an individual with ASD can interpret the emotion as fast as possible to then properly engage the individual he or she is in conversation with.    
    \item \textbf{Transient Nature of Facial Expressions}: An interesting characteristic of facial expressions is that they are highly dynamic and transient in nature \cite{3dc}, consisting of an onset, peak, and offset phase.  As a result of this transient nature of facial expressions, temporal information becomes especially important in order to capture this smooth transition as well as the transient nuances of these expressions. Traditional 2D CNNs have one major flaw in that they are primarily designed to capture spatial characteristics and not transient characteristics~\cite{CNN-RNN}. In order to encode the spatiotemporal relationships between consecutive frames, deep neural networks that leverage 3D convolutions (3DC) have been proposed, as well as architectures that leverage LSTM units to learn temporal information while avoiding the vanishing or exploding gradient problems.  However, such strategies can be very computationally complex to perform inference with, particularly since they fall within the dynamic video classification category of leveraging entire video sequences to predict emotion.
\end{itemize}
 \begin{figure*}[ht]
\includegraphics[width=\textwidth]{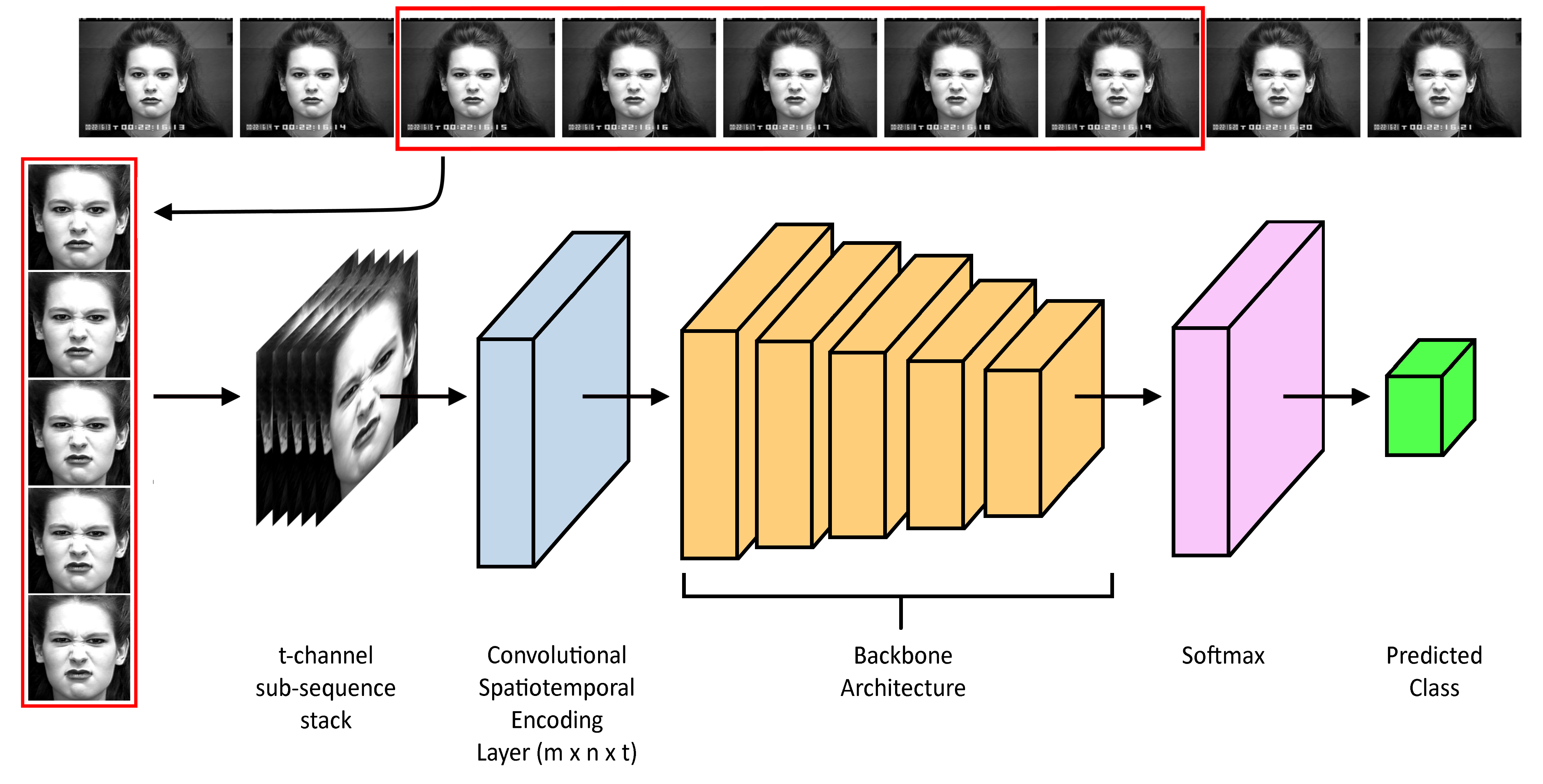}
\caption{Overview of the proposed TimeConvNet architecture. Given a streaming video sequence, a $t$-channel sub-sequence stack is constructed within a particular time window. This sub-sequence stack is then passed into a convolutional spatiotemporal encoding layer, in which learned spatiotemporal filters capture the spatial as well as the transient characteristics exhibited in the facial expressions within the sub-sequence stack. A subsequent backbone convolutional neural network architecture is then leveraged to further decompose and encode the spatiotemporal representation from the convolutional spatiotemporal encoding layer at progressively higher levels of representational abstraction, followed by a softmax layer to produce the final facial emotion classification. By performing time windowed spatiotemporal encoding, TimeConvNets strike a better balance between speed and accuracy when compared to static single image classification strategies and dynamic video classification strategies.}
\centering
\label{fig:TimeConvNet}
\end{figure*}

Motivated by the social need for assisting those with ASD to better interpret facial expressions and improve social interactions, as well as the desire to address the required low latency requirement of such systems as well as better capture the transient characteristics of facial expressions in a computationally efficient manner, this study explores a novel deep time windowed convolutional neural network design (TimeConvNets) for the purpose of real-time video facial expression recognition. More specifically, this study explores the balance between speed and accuracy by leveraging an efficient deep convolutional neural network design which perform spatiotemporal encoding of time windowed video frame sub-sequences. The applications of such a system also extend beyond the scope of assistive technology, and can also be used in many other fields, such as driver state monitoring, depression detection, analyzing customer reactions in marketing, providing additional information during security checks, ambient interfaces, and empathetic tutoring \cite{bargal2016emotion, valstar2011fully}.

Furthermore, an additional key contribution of this study is the introduction of a custom, more difficult dataset called BigFaceX through the modified aggregation of public datasets in a way that enabled time windowed learning. Famous static image datasets such as FER2013 \cite{fer2013} were unsuitable for our needs for two reasons: first, each image was completely unrelated to the others, and second, the images in the dataset show only the peak of the expression. As such, we modify and aggregate three public video datasets - the extended Cohn-Kanade (CK+) dataset \cite{CK+}, the BAUM-1 dataset \cite{baum}, and the eNTERFACE dataset \cite{enterface} to create BigFaceX in a way that facilitates for time windowed learning. Example facial expression data is shown in Fig.~\ref{fig:faces}.

The remainder of this paper is organized as follows. Section \rom{2} provides an overview of relevant literature in the field of facial expression classification. In Section \rom{3}, the proposed TimeConvNet architecture design is described and discussed in detail.  Section \rom{4} introduces the proposed custom dataset, BigFaceX, with descriptions of the incorporated datasets as well as the modified aggregation and creation process for the dataset. Experimental results for the different variants of the proposed TimeConvNet design as well as other tested networks using BigFaceX is described in Section \rom{5}, and a discussion of the experimental results is presented in Section \rom{6}, where we also discuss possible future directions of this work.  Finally, conclusions are drawn in Section \rom{7}.

\section{Related Work}

In more traditional machine learning-driven strategies for facial emotion classification, a key step in the process is the extraction of facial features from the given image or video input.  A number of handcrafted features have been proposed in research literature, with possibly the most well known being the Facial Action Coding System (FACS) \cite{Ekman1978FacialAC}. Referring to the 44 unique action units in the FACS, which can also vary in intensity based on a 5-point ordinal scale, trained human observers manually code all possible facial displays on video sequences frame by frame. Some additional examples of handcrafted features used in literature for facial emotion recognition include Gabor wavelet features \cite{littlewort2002fully}, Local Binary Patterns \cite{bargal2016emotion}, hierarchical optical flow \cite{cohn1998feature}, and Essa \& Pentland's \cite{essa1997coding} technique of extracting facial features using Principal Component Analysis (PCA) via Fast Fourier Transforms and local energy computation.  A key limitation to the use of handcrafted feature is that they may not well capture the subtle spatiotemporal nuances in facial expressions, which can be critical for high facial emotion classification performance.  Furthermore, a number of these handcrafted features can be time-consuming to calculate, thus potentially reducing the speed in which facial emotion classification can be performed. 

In recent years, there has been a lot of focus on leveraging deep learning for the purpose of facial emotion classification.  Rather than handcrafted features that could be limiting in terms of expressiveness in characterizing spatiotemporal facial nuances, deep learning approaches enable the features to be learned directly from the wealth of image and video data available.  As a result, this leads to diverse embedded features within a deep neural network that can provide a greater ability to capture such subtle yet critical facial nuances.  As a result, a wide range of deep neural network architectures such as CNNs, RNNs \cite{CNN-RNN}, and DNNs \cite{3dc} have been explored for the purpose of facial emotion classification. Additionally, authors have explored additional mechanisms to further boost the performance of deep neural networks via facial landmark injection \cite{3dc}, incorporating audio features in the network \cite{lstm, CNN-RNN}, as well as ensembling with other machine learning strategies. For example, Bargal et al. \cite{bargal2016emotion} trained three deep networks to create a concatenated feature vector that is classified using a Support Vector Machine. Pan et al. \cite{pan2019deep} used a CNN-LSTM fusion network in order to utilize temporal information in video sequences. Sun et al. \cite{lstm} performed a similar process but also included audio information, and used linear SVMs for classification. Hasani and Mahoor \cite{3dc} modified an Inception \cite{szegedy2015going} ResNet to use 3DC layers instead, followed by an LSTM unit. These studies show that the retention of temporal information can improve facial emotion classification performance by better capturing the dynamic nature of human expressions.  One limitation with the aforementioned deep learning strategies for dynamic video classification strategies is that they  generally have high computational complexities, particularly given that they leverage entire video sequences for prediction, thus making them challenging to employ in real-time, low-latency scenarios such as assisting individuals with ASD during social interactions.

\begin{figure*}[t!]
\includegraphics[width=\textwidth]{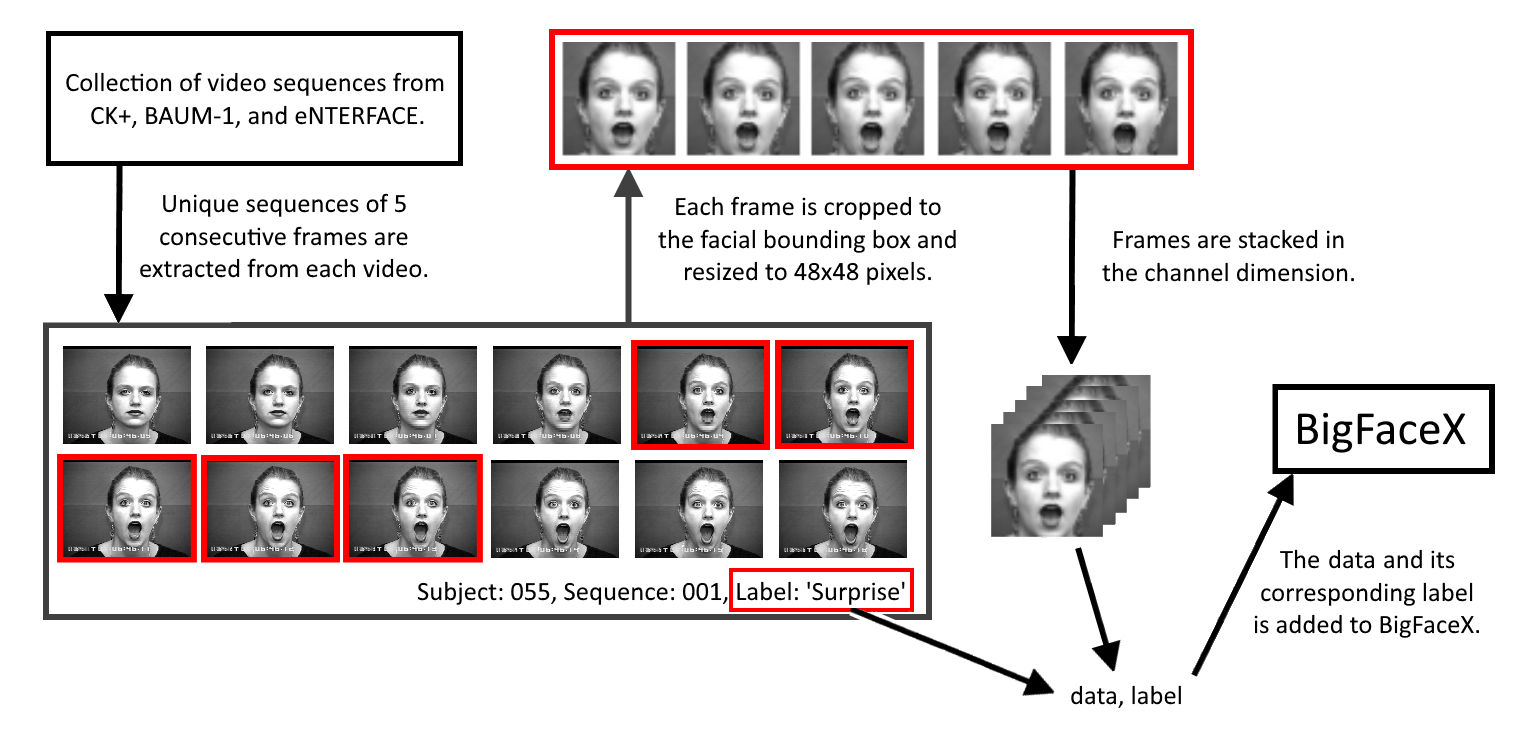}
\caption{The BigFaceX creation pipeline. Unique sequences of consecutive frames are taken from each video in CK+, BAUM-1, and eNTERFACE. Each frame is cropped to the facial bounding box, and resized. Frames are stacked together and added to BigFaceX, with the same label as the original video clip.}
\centering
\label{fig:method}
\end{figure*}

Motivated to tackle real-time facial expression recognition scenarios, a number of low latency computer vision-driven strategies have been proposed in literature.  One commonality amongst these strategies is that they tend to fit in the static single image classification category in order to reduce computational complexity.  For example, Michel \& Kaliouby \cite{michel-svm} introduced a real-time facial expression recognition system using SVMs. Various other real-time systems have also been created using various methods, such as using LBP \cite{happy2012real}, Haar feature based boosting \cite{wang2004real}, or fuzzy emotion models \cite{esau2007real}. While these low latency systems all run in real-time due to the nature of their design, they face particular difficulty as they do not capture spatiotemporal characteristics of facial expressions.  This limitation is especially prevalent  when predicting emotions of faces eliciting non-peak expressions. Thus, the ability to leverage both the high accuracy potential of spatiotemporal information as well as achieve low latency in facial emotion classification is highly desired.

In order to build machine learning driven strategies for facial emotion classification, a key ingredient is the availability of a training dataset. One of the most widely-used datasets for static single image classification is FER2013 \cite{fer2013}, which contains images of faces at the temporal peak of the expression.   Widely-used video datasets for dynamic video classification strategies include the extended Cohn-Kanade (CK+) dataset \cite{CK+}, the BAUM-1 dataset \cite{baum}, the eNTERFACE dataset \cite{enterface},  MMI dataset \cite{mmi}, and the GEMEP-FERA \cite{gemep-fera} dataset.  Another video dataset is Acted Facial Expressions in the Wild (AFEW) \cite{afew}, which is a video dataset containing facial expressions extracted from movies. A subset of this, the Static Facial Expressions in the Wild (SFEW) \cite{sfew} dataset was also made, created from extracting static frames in AFEW.  Furthermore, group emotion recognition databases also exist, such as the Happy People Images (HAPPEI) \cite{happei} database.

\section{TimeConvNet architecture}

The proposed deep time windowed convolutional neural network design (TimeConvNets) attempts to strike a balance between speed and accuracy by leveraging an  efficient design for spatiotemporal encoding of time windowed video frame sub-sequences.  As discussed previously, temporal information is a key aspect in the classification of facial expressions due to their dynamic nature. Due to this property, dynamic video classification strategies use entire video sequences in order to maximize the amount of spatiotemporal information seen at once, at the cost of high computational complexity.  By designing the proposed TimeConvNet to leverage time windowed sub-sequences, it allows us to reap the benefits of leveraging both spatial as well as transient facial cues, as dynamic video classification strategies do, while achieving significantly lower computational complexity when compared to such methods.  

The proposed TimeConvNet architecture explored in this study is shown in Figure \ref{fig:TimeConvNet}. Using a streaming video sequence as input, a $t$-channel sub-sequence stack is constructed within a specific time window.  This sub-sequence stack acts as the input to the TimeConvNet architecture, where a convolutional spatiotemporal encoding layer serves the purpose of capturing both spatial and transient characteristics within the time windowed sub-sequence stack through a set of learned convolutional filters within that layer. These learned convolutional filters, through training on time window sub-sequences of human expression video data, learns a diversity of spatiotemporal visual cues that well characterizes the different categories of emotion that we wish to predict.  A key advantage of leveraging such a time windowed convolutional encoding layer early within the architectural design is that it allows for efficiently learning spatiotemporal embeddings without the need for computationally complex 3D convolutions, as well as without needing the entire video sequence.  

In the next stage of the TimeConvNet architecture, the convolutional spatiotemporal encoding layer feeds into a backbone convolutional neural network architecture, where further hierarchical decomposition and encoding of the spatiotemporal representation from the encoding layer at progressively higher levels of representational abstraction is performed for improved discrimination amongst the categories of facial expressions. Finally, a softmax layer is used to produce the final expression prediction. Given this design, the TimeConvNet is able to leverage the transient nuances of human facial expressions alongside spatial visual cues in a computationally efficient manner that facilitates for real-time scenarios.

In this study, a number of TimeConvNet architecture variants based on different backbone architecture were constructed.  For all variants, 5-channel sub-sequence stacks were leveraged in order to reduce inference time while still providing a good characterization of transient facial behaviour. Each architecture variant was trained for 200 epochs using an initial learning rate of $1\mathrm{e}{-3}$, multiplied by $1\mathrm{e}{-1}$, $1\mathrm{e}{-2}$, $1\mathrm{e}{-3}$, and $0.5\mathrm{e}{-3}$ at epochs 81, 121, 161, and 181 respectively. Categorical cross-entropy loss was used with the Adam \cite{kingma2014adam} optimizer. Data augmentation was applied to the inputs, including rotation, width and height shifts, zoom, and horizontal flips. For training data, we leveraged the proposed BigFaceX dataset, which comprises of 68,363 data points and will be discussed in detail in the next section, and split it as follows: 70$\%$ for training, 10$\%$ for validation, and 20$\%$ for test. All experiments were run using the Intel (R) Core (TM) i3-7100 @ 3.90GHz × 4 CPU, and the GeForce RTX 2080 Ti GPU. We leveraged the Keras \cite{chollet2015keras} library for this study.

\section{BigFaceX Dataset}
A critical factor in achieving strong facial emotion classification performance lies not just in the network architecture design, but also in quality of data in which the architecture is trained on.  Motivated to establish a strong baseline dataset, we introduced the BigFaceX dataset, which is composed of a modified aggregate of three publicly available datasets: i) CK+, ii) BAUM-1, and iii) eNTERFACE. These datasets were selected as they are relatively large, and consist of video data which were ideal for extracting video sub-sequences. In total, the BigFaceX dataset contains 68,363 samples, and the class distribution is shown in Table \ref{tab:bigfacex_dist}. As shown, BigFaceX includes the six basic universal emotions, as well as a neutral class which many datasets such as FER2013 or BAUM-1 also include. Descriptions of each of the three datasets used to create BigFaceX, as well as the process employed to create BigFaceX, will be provided in the following sub-sections.

\begin{table}[htbp]
\caption{Emotion Distribution of the BigFaceX Dataset}
\begin{center}
\begin{tabular}{|c|c|}
\hline
\textbf{Emotion} & \textbf{Number of data points} \\
\hline
Angry & 8951 \\
\hline
Disgust & 8823 \\
\hline
Fear & 6069 \\
\hline
Happy & 12832 \\
\hline
Sad & 13870 \\
\hline
Surprise & 6197 \\
\hline
Neutral & 11621 \\
\Xhline{3\arrayrulewidth} % Thicker horizontal line
Total & 68363\\
\hline
\end{tabular}\par
\bigskip
\label{tab:bigfacex_dist}
\end{center}
\vspace{-0.3in}
\end{table}

%\subsection{FER2013}

%The authors of the Facial Expression Recognition 2013 dataset created it by using the Google image search API to query for images that matched a set of emotion related keywords. Images were cropped to faces, then incorrectly labelled images were manually filtered out. The resulting images were resized to 48x48 pixels, and converted to grayscale. The authors then mapped the emotion keywords into the six basic emotions plus neutral. The FER2013 dataset contains 35,887 unique static face images, and the distribution of the emotions can be found in \cite{fer2013}. This dataset is used for baseline testing due to its relevance in literature and ease of use.

\subsection{CK+}

The CK+ dataset is an improved version of the original Cohn-Kanade (CK) \cite{CK} database, with a 22 percent increase in video data and a 27 percent increase in the number of subjects. Participants ranged from 18 to 50 years of age, with a balanced mix of different genders and heritage. It contains 327 labelled emotion sequences across 123 subjects, with each sequence depicting the transition from the neutral face to the peak expression. One label is assigned to each video clip, which describes the peak expression. Each video was taken at a frame rate of 30 frames per second (FPS), with a resolution of either 640x490 or 640x480 pixels. Video clips that belong to the original CK database contained a timestamp at the bottom of each frame. As the CK+ dataset classes contain the additional emotion of contempt that is not common amongst other datasets, the samples pertaining to that emotion were not included in BigFaceX. 

\subsection{BAUM-1}

Consisting of 1,184 video clips, with 31 different subjects performing a variety of spontaneous and elicited expressions, the BAUM-1 dataset authors present it as the only non-posed database in literature at the time of publication that contains the six basic emotions in a non-English language. This dataset differs from the CK+ and eNTERFACE datasets in two major ways. First, the participants speak in Turkish rather than English, and second, it contains spontaneous reactions which are much closer to expressions in nature. Each video clip was recorded at 30 FPS, with a resolution of 720x576 pixels. 17 of the 31 subjects were female, and subjects ranged from 19-65 years of age, with a variety of hairstyles and facial hair. 3 of the subjects wore eyeglasses. This dataset also included several emotions not found in other datasets, such as the unsure, concentrating, and boredom emotions, and as such samples for those emotions are not used in BigFaceX. 

\subsection{eNTERFACE}

The eNTERFACE'05 Audio-Visual Emotion Database contained 34 male subjects and 8 female subjects from a mixture of countries. 13 of these individuals wore glasses, and 7 of them had a beard. The dataset consists of 1166 video clips, taken at a resolution of 720x576 pixels and 25 FPS. Due to the format of the data collection, all video sequences contain scripted responses, with acted expressions. One label is assigned to each video clip, which begins with a neutral expression and ends once the subject completes their scripted sentence. 

\subsection{Creating BigFaceX}

Certain pre-processing steps were required in order to create BigFaceX, which can be seen in Figure \ref{fig:method}. First, for each video sequence in the CK+, BAUM-1, and eNTERFACE datasets, a window of frame size 5 was slid across the sequence with a stride of 1 and extracted. For each window, each image was cropped to the facial bounding box using Haar Cascades \cite{viola2001rapid}, in order to remove noise from various external factors such as clothing, background, or watermarks. The bounding boxes were extended by 10$\%$ on each side, as vanilla Haar Cascades can cause the sides of the face to be removed. Each image was then resized to a size of 48x48 pixels, normalized to the range of 0 to 1, and then merged together in the channel dimension. We labelled each sub-sequence with the original label of the video it was taken from. 
Minor modifications were made when processing the BAUM-1 and eNTERFACE datasets, as they start from a neutral expression. We chose to ignore the first 5 frames of each video in order to avoid creating large amounts of mislabelled data. Furthermore, for these two datasets, a stride of 2 was used instead of 1 due to the amount of frames in the video clip. This had the effect of increasing temporal width while maintaining the small window size.

\section{Experimental Results}

To investigate the efficacy of the proposed TimeConvNet architecture design, we propose and evaluate three different TimeConvNet variants using different backbone architectures: i) mini-Xception \cite{arriaga2017real}, ii) ResNet20 \cite{resnet}, and iii) MobileNetV2 \cite{mobilenetv2}.  These three backbone architectures were chosen as they are widely used compact network architecture designs that provides a strong balance between efficiency and modeling performance. We evaluate these TimeConvNets against the following network architecture designs: i) a standard mini-Xception network architecture, ii)  a 3D ResNet20-based deep convolutional neural network architecture, and iii) a (2+1)D ResNet20-based deep convolutional neural network architecture as suggested by Tran et al. \cite{2plus1d}, who showed that this architecture is able to retain temporal relationships while lowering the number of parameters in the network when compared to 3D deep convolutional neural network architectures. Note that since the standard mini-Xception network architecture leverages a single image, only the last frame from each sub-sequence in BigFaceX was used for evaluation instead. 

The experimental results on the BigFaceX test set is shown in Table \ref{tab:results_comp}. The proposed TimeConvNet variant using the ResNet20 backbone achieved the highest top-1 accuracy (97.9$\%$) while achieving significantly lower inference time (6.14ms) that is many folds lower than the 3D ConvNet and the (2+1)D ConvNet architectures.  The fastest network architecture is the proposed TimeConvNet variant using the mini-Xception backbone, which was almost twice as fast as the ResNet20 variant (3.18ms) while still achieving higher accuracy than the the 3D ConvNet and the (2+1)D ConvNet architectures.  Furthermore, the TimeConvNet variant with mini-Xception backbone achieved  approximately 10$\%$ higher accuracy than the standard mini-Xception network architecture, while having similar inference times. The TimeConvNet variant with the MobileNetV2 backbone provided the best balance between accuracy and inference time, but possesses significantly more parameters than other networks.

\begin{table*}[t]
\caption{Comparison of tested network architectures in terms of accuracy, inference Time, and parameter count}
\begin{center}
\begin{tabular}{|c|c|c|c|}
\hline
 \textbf{Network} & \textbf{Top 1 Accuracy} & \textbf{Inf. Time (ms)} & \textbf{Parameters}\\
\hline
 2D ConvNet (mini-Xception) & 0.757 & 3.22 & \textbf{58,423} \\
\hline
 (2+1)D ConvNet (ResNet20) & 0.848 & 52.70 & 523,357 \\
\hline
3D ConvNet (ResNet20) & 0.851 & 35.64 & 808,775 \\
\Xhline{3\arrayrulewidth} % Thicker horizontal line
TimeConvNet (mini-Xception) & 0.855 & \textbf{3.18} & 58,711 \\
\hline
TimeConvNet (ResNet20) & \textbf{0.979} & 6.14 & 274,535 \\
\hline
TimeConvNet (MobileNetV2) & 0.923 & 5.64 & 2,267,527 \\
\hline
\end{tabular}\par
\bigskip
Accuracy was assessed on the BigFaceX test dataset. Inference times were averaged over 1000 runs. The best values for each column are shown in bold. 
\label{tab:results_comp}
\end{center}
\end{table*}

%The validation accuracy training curves of each model can be seen in Figure \ref{fig:val_accs}. The three TimeConvNet models has the highest accuracy, with the ResNet20 backbone model leading, followed by MobileNetV2 and then mini-Xception. The 3D ConvNet model and (2+1)D ConvNet models are next at roughly 75 percent, and finally the baseline 2D ConvNet shows the worst performance of them all. For the models that do not use mini-Xception as a backbone, there is a sudden jump in accuracy at epoch 81 which corresponds to the first learning rate reduction. We trained all models for 200 epochs each, but only show up to epoch 110 in Figure \ref{fig:val_accs} as the networks do not show significant improvement past this point even after the learning rate changes in later epochs. 

%\begin{figure}[ht!]
%\includegraphics[width=0.5\textwidth]{val_accs.png}
%\caption{Validation accuracy of the six models. Graph only shows up to epoch 110 as models do not improve further. Models without mini-Xception as a backbone show a performance jump when the learning rate is changed at epoch 81. ResNet20 has the highest final accuracy, with mini-Xception having the lowest. Best viewed in colour.}
%\centering
%\label{fig:val_accs}
%\end{figure}

\section{Discussion}

The results for the introduced TimeConvNets appear to be quite promising for the task of real-time video facial expression classification, which could potentially be very beneficial for assisting with ASD in real-time emotion recognition to improve social interactions.  Given the promising results of TimeConvNets and the BigFaceX dataset, there are a number of considerations that would be taken if leveraged within a real-time emotion recognition system.  First, it is important to consider that in addition to the time required to run inference using the network, there is also significant processing overhead associated with frame processing as well as face detection.  Based on our empirical evaluation, given that the TimeConvNets have inference times of $\sim$3-6 ms, plus additional processing overhead of around $\sim$35 ms, we can achieve a facial expression classification pipeline that takes around 40 ms per video frame, which translates to approximately 25 FPS. 

Second, it is important to note that BigFaceX, like most facial expression datasets, exhibit class imbalances, with noticeably more samples for more frequently occurring classes such as happy or sad versus less frequently encountered expressions such as fear or surprise. This can potentially lead to biases in the predictions made by the networks trained on BigFaceX.  One area of improvement in the future is a larger effort to re-balance BigFaceX. 

Third, while the BigFaceX dataset attempts to mimic real-world scenarios much better than datasets such as FER2013, as it includes non-peak expressions as well as spontaneous reactions, it is still unable to account for all possibilities, such as variation in mouth movements during speech using different languages, or varying amounts of facial occlusion. We did attempt to account for this in our design choices, such as by including the BAUM-1 dataset, as it includes non English speakers which should help the models generalize to other languages, but it is likely that a larger dataset containing languages from around the world is required. Certain amounts of facial occlusion is accounted for as datasets such as eNTERFACE include subjects with facial hair or eyeglasses, but more unique features such as piercings or tattoos may still cause inaccurate detection and prediction.  

\section{Conclusion}

The problem of facial expression recognition is important for many reasons, but especially so when trying to assist individuals with ASD. This study introduces a novel deep time windowed convolutional neural network design called TimeConvNets for the purpose of real-time video facial expression recognition, and demonstrates that by incorporating spatiotemporal encoding of time windowed video frame sub-sequences within a network architecture, as well as the use of the introduced dataset BigFaceX, TimeConvNets can achieve a good balance between accuracy, inference time, and model size.

\section*{Acknowledgment}

The authors would like to thank NSERC, Canada Research Chairs program, and Microsoft for their support.


\bibliographystyle{IEEEtran}
\bibliography{references}

\end{document}